# Predicting Cardiovascular Disease Risk using Photoplethysmography and Deep Learning

**Authors**
Wei-Hung Weng MD PhD[1†], Sebastien Baur MS[1†], Mayank Daswani PhD[1†], Christina Chen MD[1], Lauren Harrell PhD[1], Sujay Kakarmath MBBS MS[1], Mariam Jabara BS[1], Babak Behsaz PhD[1], Cory Y. McLean PhD[1], Yossi Matias PhD[1], Greg S. Corrado PhD[1], Shravya Shetty MS[1], Shruthi Prabhakara PhD[1], Yun Liu PhD[1], Goodarz Danaei ScD[2], Diego Ardila MS[1]

**Affiliations**
[1] Google LLC, Mountain View, CA, USA
[2] Department of Global Health and Population, Department of Epidemiology, Harvard School of Public Health, Boston, MA, USA
[†] W.W, S.B, and M.D. contributed equally as first authors to this work

*Address correspondence to: Wei-Hung Weng MD PhD (ckbjimmy@google.com), Shruthi Prabhakara PhD (shruthip@google.com), and Diego Ardila (ardila@google.com)

## Abstract

Cardiovascular diseases (CVDs) are responsible for a large proportion of premature deaths in low- and middle-income countries. Early CVD detection and intervention is critical in these populations, yet many existing CVD risk scores require a physical examination or lab measurements, which can be challenging in such health systems due to limited accessibility. Here we investigated the potential to use photoplethysmography (PPG), a sensing technology available on most smartphones that can potentially enable large-scale screening at low cost, for CVD risk prediction. We developed a deep learning PPG-based CVD risk score (DLS) to predict the probability of having major adverse cardiovascular events (MACE: non-fatal myocardial infarction, stroke, and cardiovascular death) within ten years, given only age, sex, smoking status and PPG as predictors. We compared the DLS with the office-based refit-WHO score, which adopts the shared predictors from WHO and Globorisk scores (age, sex, smoking status, height, weight and systolic blood pressure) but refitted on the UK Biobank (UKB) cohort. In UKB cohort, DLS's C-statistic (71.1%, 95% CI 69.9-72.4) was non-inferior to office-based refit-WHO score (70.9%, 95% CI 69.7-72.2; non-inferiority margin of 2.5%, p<0.01). The calibration of the DLS was satisfactory, with a 1.8% mean absolute calibration error. Adding DLS features to the office-based score increased the C-statistic by 1.0% (95% CI 0.6-1.4). DLS predicts ten-year MACE risk comparable with the office-based refit-WHO score. It provides a proof-of-concept and suggests the potential of a PPG-based approach strategies for community-based primary prevention in resource-limited regions.

## Introduction

Cardiovascular diseases (CVDs) are responsible for one third of deaths globally [1] with approximately three quarters occurring in low- and middle-income countries (LMICs) where there's a paucity of resources for early disease detection [2,3]. Because CVD risk factors such as hypertension, diabetes, or hyperlipidemia are typically symptomless before advanced disease, there is a great need for screening programs to identify those at high risk of CVD events. Interventions such as lifestyle counseling, with or without prescription medications, have shown to be an effective strategy for CVD prevention among these individuals [4].

Multiple risk scores, such as WHO/ISH risk chart and Globorisk scores, have been developed to triage CVD risk based on demographics, past medical history, vital signs, and laboratory data [4–7]. However, the dependency of these risk scores on medical and laboratory equipment (e.g., sphygmomanometers) [8,9] limits their reach. Specifically, low-resource healthcare systems have relied largely on opportunistic screening [10], such as via community healthcare workers (CHWs) [11], to close access gaps. We reasoned that developing low-cost, easy-to-use, lightweight, digital point-of-care tools using sensors already available in smartphones [12–14], could potentially further the reach and capability of CHW-based programs and enable large-scale screening at low cost [15].

Among sensing signals for the circulatory system, photoplethysmography (PPG) is a non-invasive, fast, simple, and low-cost technology, and can be captured with sensors available on increasingly ubiquitous devices such as smartphones and pulse oximeters [16]. PPG measures the change in blood volume in an area of tissue across cardiac cycles and is primarily used for heart rate monitoring in healthcare settings [17,18]. Research has also investigated the utility of PPG in understanding short term fluctuations in vascular compliance, by estimating continuous blood pressure (BP) in an ICU setting [17,19,20], though the accuracy of such approaches is known to be insufficient even when per-user calibration is available [17]. Beyond short term vascular changes, research has also been conducted into understanding the slow manifestation of vascular aging and arterial stiffness from PPG waveforms [21–23], which are useful for longer-term CVD risk assessment. Since PPG is potentially more accessible and requires less training for measurement, such technologies could provide accurate real-time insights. The ubiquity of smartphones have also prompted research involving PPG as measured from smartphone cameras, via placing a finger on the camera [16]. Taken together, enabling CVD-risk estimation based on PPG signals can potentially be a highly accessible screening tool in low-resource health systems (Figure 1a).

In this paper, we investigate the feasibility of leveraging PPG for CVD risk prediction using data from the UK Biobank (UKB). Specifically, we predict the ten-year risk of developing a major adverse cardiovascular event (MACE) using deep learning-based PPG embeddings and heart rate (measured by PPG), along with other demographics, including age, sex and smoking status, but without any inputs from physical examination or laboratory data (Supplementary Figure 1). We find that our deep learning PPG-based CVD risk prediction score (DLS) is well-calibrated and non-inferior to the existing comparative office-based CVD risk score using predictors from WHO/ISH and Globorisk, that requires blood pressure, weight and height measurement, or laboratory data.

## Results

We showed that DLS demonstrated non-inferiority to the office-based refit-WHO score. We evaluated the ten-year MACE risk prediction performance of all methods using our UKB test subset, which was held-out during the training process. The DLS yielded C-statistic of 71.1% (95% CI [69.9, 72.4]). When compared with the office-based refit-WHO score, the DLS was non-inferior ($p<0.01$), with a delta of +0.2% (-0.4, 0.8). The cfNRI was 0.1% (0.0, 0.1), stemming primarily from improved reclassification of events (0.1% [0.0, 0.2]), without performance penalty in the non-events (0.0% [0.0, 0.0]).

Based on the C-statistic, there was an incremental improvement when the metadata model (69.1%) was augmented with manually engineered (not deep learning derived) PPG morphology features (70.0%). The DLS was superior to this metadata+PPG morphology features model ($p<0.01$), indicating value in deep learning based feature extraction. The lab-based model (which requires total cholesterol and glucose information) was superior to the office-based refit-WHO score (71.6% versus 70.9%, $p<0.01$). By applying the Globorisk scores in [7], which recalibrating on UKB cohort for baseline hazard and mean risk factors but without re-estimating the coefficients, the office-based Globorisk yielded a C-statistic of 70.0% (68.8, 71.2), and the lab-based Globorisk yielded a C-statistic of 69.8% (68.5, 71.1). More details are shown in Table 1.

For a fair comparison, we then selected the risk thresholds that matched the specificity or sensitivity of SBP-140 (see Statistical analysis in Methods) (specificity of 63.7%, sensitivity of 55.2%). We found that at matched specificity, the sensitivity of the DLS (67.9%) were non-inferior to the office-based refit-WHO score (67.7%) (p=0.012), and a comparable NRI, while the metadata and metadata + PPG morphology models were not (p=0.984 and p=0.305, respectively). At the matched sensitivity, the DLS's specificity (74.0%) was also non-inferior to the baseline (73.1%) (showed superiority with p<0.01), with a comparable NRI. The laboratory-based refit-WHO and the model using metadata and PPG morphology-based features were also non-inferior to the office-based refit-WHO score, despite these models requiring additional inputs from laboratory measurements or engineered PPG features, respectively. The metadata-only model performed more poorly than the office-based refit-WHO score across different metrics. We also conducted Kaplan Meier analysis on risk groups defined using the above approach (Figure 1b). Both thresholds showed significant (p<0.01, log rank tests) differences between the groups. Results for the 10% risk threshold are in Supplementary Table 1 and Supplementary Figure 2.

In addition to the default set of inputs to the DLS, we also evaluated models with BMI, and with both BMI and SBP (both of which are predictors in the office-based refit-WHO) included as additional inputs, which we refer to as DLS+ and DLS++, respectively. We found that adding BMI (DLS+) improved DLS in terms of both discrimination and net reclassification. Additional improvement was observed after adding SBP (DLS++), which further improved the DLS model, and demonstrated superiority across different metrics (Supplementary Table 2a, 2b). We also showed that for DLS and its variants (DLS+ and DLS++), the cfNRI and NRI with different risk thresholds (Supplementary Table 2a) were also on par with the office-based refit-WHO score (Supplementary Table 2b). These findings indicate that combining the existing non-laboratory risk factors from the refit-WHO score with the DLS features yields a more accurate CV risk estimation. We further developed a model (Full model) that includes more risk factors used in the CVD risk scores commonly used in high-income countries (QRISK and/or ASCVD), as well as a model that incorporates genetic risk, and listed the findings in the Supplementary Results.

We also examined the association between each model and MACE via the coefficients and hazard ratios (HRs) (Supplementary Table 3). We found that in the office-based refit-WHO score, smoking, older age, higher BMI, and higher SBP were associated with the ten-year MACE risk. We also found that some DLS features were also associated with the ten-year MACE risk ($p<0.05$ for four deep learning PPG features in DLS and DLS+, and for two PPG features for DLS++).

Meanwhile, the predicted and observed risks of ten-year MACE were similar across different models (Figure 1c), which indicated DLS has similar calibration performance compared with other models. The calibration slope of DLS was similar to the office-based refit-WHO score (0.981 versus 0.979) (Table 2). We also found that DLS++ has a comparable calibration performance (Supplementary Table 2a, 2b). All models except the DLS+ have an observed ten-year MACE risk estimation within 5% mean absolute calibration error (i.e., the slopes were between 0.95-1.05).

Finally, DLS is on par with the office-based refit-WHO score in some subgroups. Supplementary Table 4 shows that DLS demonstrated non-inferiority in some subgroups and showed superiority in smoking, hypertensive and male subgroups. Both the office-based refit-WHO score and DLS had similar performance trends. Both models have higher sensitivity and lower calibration error but lower specificity on the smoking, older, male, and hypertensive subgroups. The models were well-calibrated for most subgroups, but systematically overestimated absolute risk about 4.0% in the elevated A1c and about 1.0% in hypertensive subgroups. The finding indicates that the developed risk models tend to be better calibrated and better predict ten-year MACE risk in a population that has higher known CVD risk factors, such as older, male, smoking, higher blood glucose and hypertensive subgroups (Supplementary Table 4). Across different age, sex, smoking, and comorbidity (diabetes and hypertension) subgroups, the calibration for all risk scores were similar in predicting ten-year MACE risk in smoking, age greater than 55, male, not elevated A1c populations, with prediction errors within 10% (i.e. the calibration regression slope between 0.9 and 1.1 (Supplementary Table 4, Supplementary Figure 3)).

## Discussion

We developed a deep learning PPG-based CVD risk score, DLS, to predict ten-year MACE risk using age, sex, smoking status, heart rate and deep learning-derived PPG features. Without requiring any vital signs or laboratory measurement, DLS demonstrated non-inferior performance compared to the office-based refit-WHO score with coefficients re-estimated on the same cohort. Results were consistent between metrics (C-statistic, NRI, cfNRI, sensitivity, specificity, calibration slope), and in various subgroups. Improved cfNRI and NRI also indicate the capability of DLS to reclassify cases better than the office-based refit-WHO score. Additionally, if available, adding office-based features (BMI, SBP) on top of DLS further improved the model performance.

Our work focuses on understanding the role that PPG and deep learning can play in settings where equipment access to healthcare is limited, such as community-based screening programs in LMICs. Several CVD prediction scores without an assumption of the availability of laboratory measurement exist for primary prevention, such as WHO/ISH risk prediction chart [24], office-based Framingham risk score (FRS) [25], office-based Globorisk score [4], non-laboratory INTERHEART risk score [26], and Harvard NHANES risk score [27]. Some of these are also deployed in real-world clinical practice [4,28], though these methods require either body measurements (BMI, waist-hip ratio), SBP, or both. Challenges remain in scaling up CVD screening in the resource-limited areas due to reasons such as the lack of laboratory devices, sphygmomanometer cuffs, or the necessary training of CHWs for accurate measurements. In our study, the DLS demonstrated performance comparable to that of the re-estimated office-based refit-WHO score, without requiring accurate laboratory examination, vital signs measured via additional devices, or BMI. This feature improves accessibility for health systems that have limited resources to collect vitals and labs for CVD risk screening and triage. More intriguing, PPG signals could in principle be captured through a smartphone [16], and future work could leverage smartphone-based PPGs along with the DLS to enable large-scale screening and triage in the community at low cost (Figure 1a) [14,29].

Due to the higher prevalence, lower diagnosis rate and lower treatment of CVD in LMICs, WHO has listed preventing and controlling CVD as main targets in their "Global action plan for the prevention and control of non-communicable diseases (NCDs) 2013-2030" [30]. PPG-based screening may allow healthcare systems to optimize use of resources by funneling in those who are likely to benefit the most and improve the early detection of CVDs. Thus, our study represents a step on the journey towards enabling community-based preventive treatment for high CVD risk individuals with limited healthcare access.

The deep learning-based features are challenging to interpret directly, and the pathophysiology between PPG and CVD risk is still under investigation [31]. We computed the Pearson correlation coefficient between DLS features and engineered PPG morphological features (Supplementary Table 5), and found some correlations exceeded 0.3. We also found that using summarized resting electrocardiogram (ECG) yielded a comparable performance against the office-based model on a UKB subset containing resting ECG, with the C-statistic of 70.9% (56.9, 83.0) versus 69.9% (56.7, 81.1) (p=0.845), yet further evidence is required to draw conclusions.

Several limitations of the study should also be noted. We used a single dataset, UKB, for both modeling and evaluation. Though we have stratified the UKB cohort based on geographical information to allow for non-random variation [32], further work is needed to understand generalization to other populations. Notably, UKB is not representative of the population in LMICs. However, using UKB to demonstrate the capability of using DLS for long-term CVD risk prediction is an important first step in justifying a prospective data collection in LMICs. The device used for PPG acquisition in the UKB is a clinical pulse oximeter, thus our results provide direct evidence that the pulse oximeter may be a reliable CVD screening tool. Studies have found that the heart rate and rhythm extracted from smartphone PPG were comparable with clinical grade devices such as ECG [33–35], but additional work is needed to know if deep learning models can be developed directly on smartphone-collected PPGs. Future work could focus on predicting CVD risk using prospective smartphone PPG datasets from low-resource healthcare systems.

To summarize, our study found that a deep learning model extracted features that when added to easily extractable clinical and demographic variables (such as smoking status, age and sex), provided statistically significant prognostic information about cardiovascular risk. Our work is an initial step towards accurate and scalable CVD screening in resource-limited areas around the world.

## Methods

### Overview

We developed a new CVD risk prediction score, DLS, using age, sex, smoking status and the results of analysis of PPG signals using deep learning. We used a Cox proportional hazard model and data from UKB to predict the ten-year risk of MACE among individuals free of CVD at baseline.

### Data Source and Cohort

The DLS was developed and evaluated using data from the UKB dataset, filtered to focus on participants aged 40-74 to mirror a previous study [4]. We then stratified UKB participants who had PPG waveforms recorded into three subsets: train (n=105,319), tune (n=46,868), and test (n=57,702) subsets based on geographic information on the site of data collection, i.e., latitude and longitude. This strategy aligns with TRIPOD guidelines [32] on external validation (specifically validation on a different geographic region) by allowing for non-random variation between data splits such as differences in data acquisition or environment.

We used PPG waveforms from all visits for the participants in this train subset to train the PPG feature extractor in DLS (details in "Model Development"). The low-dimensional numeric outputs (embeddings) computed by this model were used as additional input features to our Cox model. To develop the Cox model that generates DLS to predict MACE risk, additional clinical and demographic variables and inclusion/exclusion criteria were needed. First, we excluded participants with non-fatal myocardial infarction or stroke before their first visit, or missing any of variables for our model (age, sex, and smoking status). We also excluded those without body mass index (BMI) or systolic BP (SBP) for a fair comparison against the other office- and lab-based risk prediction models. For each participant, we only included measurements related to their first visit. All numerically measured variables were standard-scaled. Cox models were regularized using a ridge penalty. In the final cohort, 97,970, 43,539, and 54,856 participants were included to train, tune, and test the survival model, respectively (Figure 2). Descriptive statistics for this cohort are in Table 3.

## Model Development

### First Stage: PPG Feature Extractor

For DLS, we first trained a deep learning-based feature extractor to learn PPG representations from raw PPG waveform signals, using a one-dimensional ResNet18 [21,22,36] as the neural network architecture. We trained the feature extractor on the train subset, and picked the network weights that maximized the Cox pseudolikelihood (see description of the second stage below) on the tune subset. These weights were used to compute PPG embeddings on the train, tune, and test subsets. The PPG embeddings were further processed by principal component analysis (PCA)–a technique to reduce dimensionality–to five PCA-derived DLS features that are used by the survival model. Modeling details are listed in Supplementary Methods.

### Second Stage: Survival Model

In the second stage, we developed a Cox proportional hazards regression model for predicting ten-year MACE risk, using as inputs age, sex, smoking status, PCA-derived PPG embeddings and PPG-HR (heart rate measured during PPG assessment). The model was trained on the train subset and tuned on the tune subset to decide the best-performing ridge regularization parameter (Supplementary Table 6).

### Models for Comparison/Reference

For comparisons, we developed different survival models based on different feature sets (Table 3 "Features used" column, and Supplementary Table 7), including office-based and laboratory-based refit-WHO scores using the CVD risk predictors adopted in WHO/ISH risk chart and Globorisk studies, metadata-only model (age, sex, smoking status), metadata + PPG morphology (a model with metadata and engineered PPG features describing waveform morphology, such as dicrotic notch presence, details in Supplementary Table 5), a model without smoking status as an input (metadata without smoking, DLS without smoking), and the "Full" model that considered metadata, laboratory data, medication and medical history as a reference, to compute CVD risk score. We chose the model using shared predictors from the office-based WHO/ISH risk chart and Globorisk score (office-based refit-WHO score) as the main reference since they are adopted in the CVD risk research for low-resource settings. To ensure the fairest comparison the coefficients for the WHO and Globorisk predictors were re-fitted using the same UKB train subset as our DLS, and a sensitivity analysis was conducted using the original coefficients with recalibration.

We further developed DLS+ (DLS with BMI), and DLS++ (DLS with BMI and SBP) that additionally included more non-laboratory, office-based measurements as inputs of the survival model to better understand the prognostic value of PPG on top of the existing office-based refit-WHO model.

All models were trained on the same train subset and tuned on the tune subset except for laboratory-based refit-WHO score, metadata + PPG morphology, and the Full models that we trained, tuned and compared based on a subset of the testing data without missing values of the input features.

## Evaluation

### Endpoints

The outcome, ten-year risk of MACE, was defined as a composite outcome of three components, non-fatal myocardial infarction, stroke, and CVD-related death (using ICD codes and cause of death to identify, Supplementary Table 5 for details) [7,28]. To define the outcome, we used (1) the date of heart attack, myocardial infarction, stroke, ischemic stroke, either diagnosed by doctor or self-reported, (2) the record of ICD-10 (international classification of diseases, 10th revision) clinical codes, and (3) and strings that are associated with the CVD-related death. The ICD-10 codes used included I21 (acute myocardial infarction), I22 (subsequent myocardial infarction), I23 (complications after myocardial infarction), I63 (cerebral infarction), I64 (stroke not specified as hemorrhage or infarction). The strings we used for matching include those related to coronary artery diseases, myocardial infarction, stroke, hypertensive diseases, heart failure, thromboembolism, arrhythmia, valvular diseases and other heart problems. We used the earliest date on any of the data sources mentioned above as the outcome date.

## Statistical analysis

For primary analysis, we compared DLS with the office-based refit-WHO score, which is a risk model for healthy individuals across different countries [4,7,24,37], using Harrell's C-statistic. We conducted a non-inferiority test with a pre-specified margin of 2.5% and alpha of 0.05, both selected based on power simulations using the tune subset. For secondary analyses, we also compared DLS with scores generated by other models mentioned in "Models for Comparison/Reference" above.

Additional evaluation metrics included the category-free net reclassification improvement (cfNRI) [38], and after defining a specific risk threshold (model operating point), sensitivity, specificity, NRI, and adjusted hazard ratio (HRs). For NRI and cfNRI, we also reported the respective event and non-event components. Risk thresholds were selected in three ways: (1) matching the sensitivity of SBP-140 (described next), (2) matching the specificity of SBP-140, and (3) the 10% predicted risk threshold suggested by the Globorisk study [4]. Elevated SBP above 140 mmHg ("SBP-140") [39] was used for threshold selection because it is used as a simple single-visit indicator of BP control in the healthcare program of some countries such as India [40], and we hypothesized that the PPG provided a single-visit indicator of vascular properties. To calculate sensitivity and specificity, we excluded the participants without a ten-year follow up if they didn't have a MACE event within ten years. To evaluate model calibration, we used the slope of the line comparing predicted and actual event rates, for deciles of model prediction [37]. We also performed subgroup analyses based on smoking status, sex, age, elevated HbA1c and hypertension status. We used quintiles for the elevated HbA1c subgroup due to the smaller sample size.

For statistical precision, we used the Clopper-Pearson exact method to compute the 95% confidence intervals (CIs) for sensitivity and specificity, and used the non-parametric bootstrap method with 1,000 iterations to compute 95% of all remaining metrics and delta values. For hypothesis tests in secondary and exploratory analysis, we used a permutation test to examine the non-inferiority and superiority of the C-statistic, and the one-sided Wald test for sensitivity and specificity. The log-rank test was used to determine whether survival differs between the model-defined low and high risk groups. For all two-sided tests, we used an alpha value of 0.05.

## Acknowledgements

This work was funded by Google LLC. All authors affiliated with Google are employees and own stock as part of the standard employee compensation package. We acknowledge Nick Furlotte (Google Research) and the Google Research team for software infrastructure support. We also thank Boris Babenko (Google Research) for his critical feedback on the manuscript. This research was conducted with the UK Biobank resource application 65275.

## Author contributions

WHW, SB, MD contributed equally to the study. WHW, SB, MD, SK, SP, DA conceived the study. WHW, SB, MD, YL designed the methodology and protocol. SB, MD, MJ managed the study, data collection, model training and infrastructure. WHW, SB, MD, CC, BB, CYM analyzed and interpreted the data. WHW, MD, LH conducted statistical analysis. WHW drafted the manuscript. SB, MD, CC, SK, YL, GD, DA critically reviewed and edited the manuscript for important intellectual content. WHW, SB, MD, DA verified the data. YM, GSC, SS, SP obtained research funding and resource support. SK, SP, GD provided administrative and technical support. WHW, SP, DA supervised the research. All authors had access to all the data in the study and had final responsibility for the decision to submit for publication.

## Data availability

This research has been conducted using the UK Biobank Resource under Application Number 65275. Please visit the UK Biobank website, https://www.ukbiobank.ac.uk/, for data application procedures.

## Code availability

The deep learning framework (JAX) used in this study is available at https://github.com/google/jax [41]. All survival analyses were conducted using Lifelines [42], an open source Python library.

## Declaration of competing interests

Author WHW, SB, MD, CC, YL, and DA are employed at Google LLC and hold shares in Alphabet, and are co-inventors on patents (in various stages) for CVD risk prediction using deep learning and PPG, but declare no non-financial competing interests. LH, BB, CYM, YM, GSC, SS, SP are employed at Google LLC and hold shares in Alphabet but declare no non-financial competing interests. SK did this work at Google via PRO Unlimited, holds shares in Alphabet, serves as an Associate Editor for this journal but had no role to play in the editorial process and decisions for this manuscript, and is a co-inventor on patents for CVD risk prediction using deep learning and PPG. GD declares no financial or non-financial competing interests.

# Tables

**Table 1: Model performance comparison of 10-year major adverse cardiovascular events (MACE) risk prediction between DLS versus other methods for the non-operating point dependent metrics.** The primary analysis of the study is non-inferiority of the C-statistic of the DLS model compared with the office-based refit-WHO model. 95% confidence intervals (CIs) of C-statistic, cfNRI, and slope were obtained via bootstrapping, and the p-values were computed via a permutation test. The slope was not calculated for SBP-140 since its output is binary. *In the "Feature used" column, "Metadata" includes age, sex, and smoking status.

| **Model** | **C-statistic (%)** | **Delta (%)** | **P-value for non-inferiority of C-statistic** | **P-value for superiority of C-statistic** | **cfNRI (%)** | **cfNRI (event) (%)** | **cfNRI (non-event) (%)** | **Calibration slope** | **Features used*** |
|---|---|---|---|---|---|---|---|---|---|
| Office-based refit-WHO | 70.9 (69.7, 72.2) | n/a (reference) | | | | | | 0.979 (0.915, 1.038) | Metadata + BMI + SBP |
| DLS | 71.1 (69.9, 72.4) | 0.2 (-0.4, 0.8) | <0.01 | 0.292 | 0.1 (0.0, 0.1) | 0.1 (0.0, 0.2) | 0.0 (0.0, 0.0) | 0.981 (0.919, 1.045) | Metadata + PPG |
| Metadata | 69.1 (67.9, 70.4) | -1.7 (-2.2, -1.3) | <0.01 | 1 | -0.4 (-0.5, -0.3) | -0.2 (-0.3, -0.2) | 0.2 (0.1, 0.2) | 0.94 (0.875, 1.004) | Metadata |
| SBP-140 | 59.4 (58.3, 60.5) | -11.5 (-12.7, -10.2) | 1 | 1 | -1.3 (-1.4, -1.2) | -1.0 (-1.1, -0.9) | 0.3 (0.3, 0.3) | - | SBP |
| Evaluated on the subset with all PPG morphology data available | | | | | | | | | |
| Metadata + PPG morphology | 70.0 (68.8, 71.3) | -0.9 (-1.4, -0.4) | <0.01 | 1 | -0.1 (-0.2, -0.1) | -0.1 (-0.1, 0.0) | 0.1 (0.1, 0.1) | 1.02 (0.951, 1.086) | Metadata + engineered PPG features |
| Office-based refit-WHO | 70.9 (69.7, 72.2) | n/a (subset reference) | | | | | | 0.977 (0.913, 1.035) | (Metadata + BMI + SBP) |
| Evaluated on the subset with laboratory data available | | | | | | | | | |
| Lab-based refit-WHO | 71.6 (70.4, 72.9) | 0.5 (0.1, 0.9) | <0.01 | <0.01 | 0.2 (0.1, 0.2) | 0.3 (0.2, 0.3) | 0.1 (0.1, 0.1) | 0.921 (0.864, 0.982) | Metadata + total cholesterol + glucose |
| Office-based refit-WHO | 71.1 (69.9, 72.4) | n/a (subset reference) | | | | | | 0.897 (0.838, 0.959) | (Metadata + BMI + SBP) |

**Table 2: Model performance comparison of 10-year major adverse cardiovascular events (MACE) risk prediction between DLS versus each of other methods using a risk threshold matches the same specificity or sensitivity of SBP-140.** 95% confidence intervals (CIs) of sensitivity and specificity were obtained from the Clopper-Pearson exact method, and the p-values were calculated by the permutation test with the prespecified margin of 2.5% and alpha of 0.05. The 95% CIs of NRI were computed via bootstrapping.

| | **Sensitivity@specificity of 63.7%** | | | | | | | **Specificity@sensitivity of 55.2%** | | | | | | |
|---|---|---|---|---|---|---|---|---|---|---|---|---|---|---|
| **Model** | **Mean (%)** | **Delta (%)** | **Non-inferiority p-value** | **Superiority p-value** | **NRI (%)** | **NRI (event) (%)** | **NRI (non-event) (%)** | **Mean (%)** | **Delta (%)** | **Non-inferiority p-value** | **Superiority p-value** | **NRI (%)** | **NRI (event) (%)** | **NRI (non-event) (%)** |
| Office-based refit-WHO | 67.7 (65.2, 70.1) | reference | | | | | | 73.1 (72.7, 73.5) | reference | | | | | |
| DLS | 67.9 (65.4, 70.3) | 0.1 (-1.9, 2.0) | <0.01 | 0.654 | -0.3 (-2.0, 1.6) | 1.0 (-0.9, 2.9) | 1.2 (0.9, 1.5) | 74.0 (73.6, 74.4) | 0.9 (-0.7, 2.5) | <0.01 | <0.01 | 1.1 (-0.9, 3.1) | 1.6 (-0.5, 3.4) | 0.4 (0.1, 0.8) |
| Metadata | 63.6 (61.0, 66.0) | -4.1 (-5.9, -2.2) | 0.984 | 1 | -2.7 (-4.5, -0.9) | 0.8 (-0.9, 2.6) | 3.5 (2.9, 3.8) | 70.4 (70.0, 70.8) | -2.4 (-3.9, -1.1) | 0.961 | 1 | -2.4 (-4.9, -0.3) | -0.4 (-3.1, 1.6) | 2.0 (0.6, 2.7) |
| SBP-140 | 55.4 (53.1, 57.9) | -12.4 (-15.1, -9.3) | 1 | 1 | -11.1 (-13.6, -8.2) | -18.7 (-21.2, -15.8) | -7.7 (-8.1, -7.2) | 63.6 (63.2, 64.0) | -11.0 (-74.4, -8.5) | 1 | 1 | -6.8 (-31.6, -3.1) | 12.6 (8.9, 27.8) | 16.6 (16.0, 56.5) |
| Evaluated on the subset with all PPG morphology data available | | | | | | | | | | | | | | |
| Metadata + PPG morphology | 66.0 (63.5, 68.5) | -1.8 (-3.5, -0.1) | 0.305 | 0.992 | -1.6 (-3.4, 0.2) | 0.3 (-1.5, 2.2) | 1.9 (1.7, 2.2) | 71.7 (71.3, 72.0) | -1.5 (-2.9, -0.1) | <0.01 | 1 | -1.5 (-3.7, 0.7) | -0.2 (-2.4, 2.0) | 1.3 (1.1, 1.6) |
| Office-based refit-WHO | 67.7 (65.3, 70.1) | n/a (subset reference) | | | | | | 73.1 (72.8, 73.5) | n/a (subset reference) | | | | | |
| Evaluated on the subset with laboratory data available | | | | | | | | | | | | | | |
| Lab-based refit-WHO | 69.1 (66.6, 71.6) | 1.0 (-0.7, 2.5) | <0.01 | 0.106 | 2.0 (0.3, 3.7) | 4.2 (2.5, 5.8) | 2.2 (1.9, 2.4) | 74.8 (74.4, 75.2) | 1.5 (0.1, 2.5) | <0.01 | <0.01 | 2.7 (1.1, 4.4) | 4.3 (2.6, 6.0) | 1.5 (1.3, 1.8) |
| Office-based refit-WHO | 68.2 (65.7, 70.7) | n/a (subset reference) | | | | | | 73.4 (73.0, 73.8) | n/a (subset reference) | | | | | |

**Table 3: Cohort statistics for 10-year major adverse cardiovascular events (MACE) risk prediction at the first UK Biobank visit.**

| | **Overall** | **Train** | **Tune** | **Test** |
|---|---|---|---|---|
| Number of sites | 15 | 9 | 3 | 3 |
| Geographic location of sites | See train/tune/test | Swansea, Bristol, Birmingham, Nottingham, Sheffield, Cheadle, Wrexham | Newcastle, Middlesborough, Liverpool | Croydon, Hounslow, Reading |
| Patients | 196,365 | 97,970 | 43,539 | 54,856 |
| MACE, count (%) | 5,650 (2.9) | 2,798 (2.9) | 1,401 (3.2) | 1,451 (2.6) |
| Age, median [IQR] | 59.0 [51.0,64.0] | 59.0 [52.0,65.0] | 60.0 [52.0,64.0] | 58.0 [50.0,63.0] |
| Sex=female, count (%) | 107,679 (54.8) | 52,885 (54.0) | 23,833 (54.7) | 30,961 (56.4) |
| Smoking status, count (%) | 84,111 (42.8) | 41,559 (42.4) | 18,436 (42.3) | 24,116 (44.0) |
| BMI, median [IQR] | 26.6 [24.0,29.7] | 26.6 [24.1,29.7] | 26.9 [24.3,30.0] | 26.2 [23.7,29.4] |
| SBP, median [IQR] | 136.5 [124.5,149.5] | 137.0 [125.5,150.0] | 139.0 [127.0,152.0] | 133.5 [122.0,146.5] |
| Total cholesterol, median [IQR] | 5.7 [5.0,6.5] | 5.7 [5.0,6.5] | 5.7 [5.0,6.5] | 5.6 [4.9,6.4] |
| HDL, median [IQR] | 1.4 [1.2,1.7] | 1.4 [1.2,1.7] | 1.4 [1.2,1.7] | 1.4 [1.2,1.7] |
| HbA1c, median [IQR] | 35.3 [32.9,38.0] | 35.3 [32.9,37.9] | 35.3 [32.9,37.9] | 35.4 [32.9,38.2] |
| Diabetes, count (%) | 4738 (2.4) | 2129 (2.2) | 1063 (2.4) | 1546 (2.8) |
| Hypertension, count (%) | 52,299 (26.6) | 26,097 (26.6) | 11,618 (26.7) | 14,584 (26.6) |
| History of angina, count (%) | 161 (0.1) | 89 (0.1) | 34 (0.1) | 38 (0.1) |
| History of hyperlipidemia, count (%) | 18,872 (9.6) | 8,517 (8.7) | 4,380 (10.1) | 5,975 (10.9) |
| On hypertension medication, median [IQR] | 39762 (20.2) | 20072 (20.5) | 9151 (21.0) | 10539 (19.2) |
| On statin, median [IQR] | 32297 (16.4) | 16068 (16.4) | 7431 (17.1) | 8798 (16.0) |

# Figures

**Figure 1: Summary of study motivation and design and main results.** (a) The motivation of applying the PPG-based cardiovascular disease (CVD) risk assessment in the low-resource health systems. Non-office based information acquired from mobile-sensing technologies may help address the burden of cardiovascular disease risk screening and triage in resource-limited areas. (b) Kaplan-Meier curves for the DLS with different definitions of high risk. Left: risk threshold corresponding to a specificity of 63.6%, right: risk threshold corresponding to a sensitivity of 55.4% (see Methods). The p-values were calculated by the log-rank test. (c) Calibration plot, showing observed and predicted 10-year MACE risk. We discretized each model's output into deciles and the slopes indicate the coefficient of a linear regression.

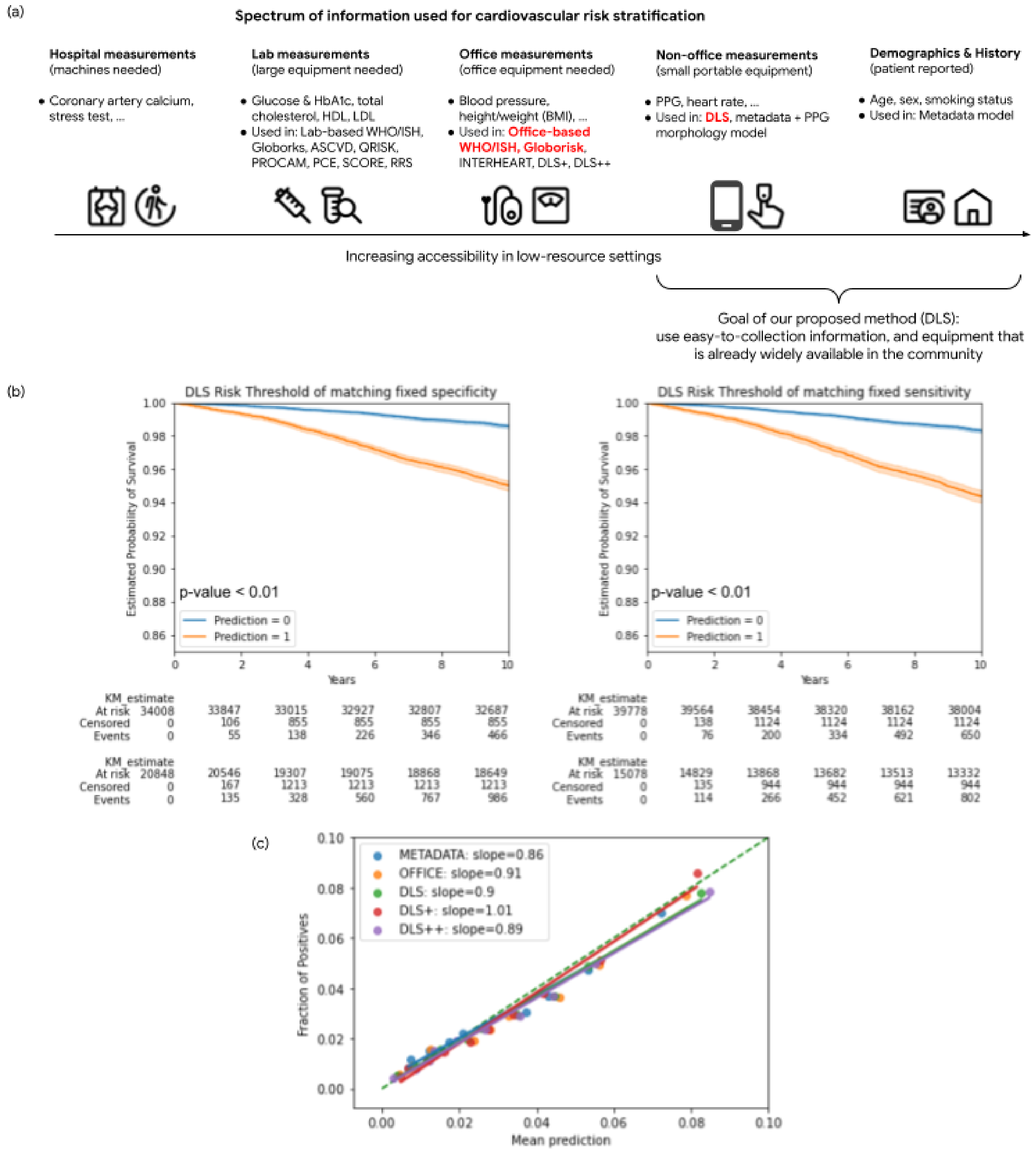

**Figure 2: Flowchart of inclusion and exclusion of the cohort for developing the survival model.**

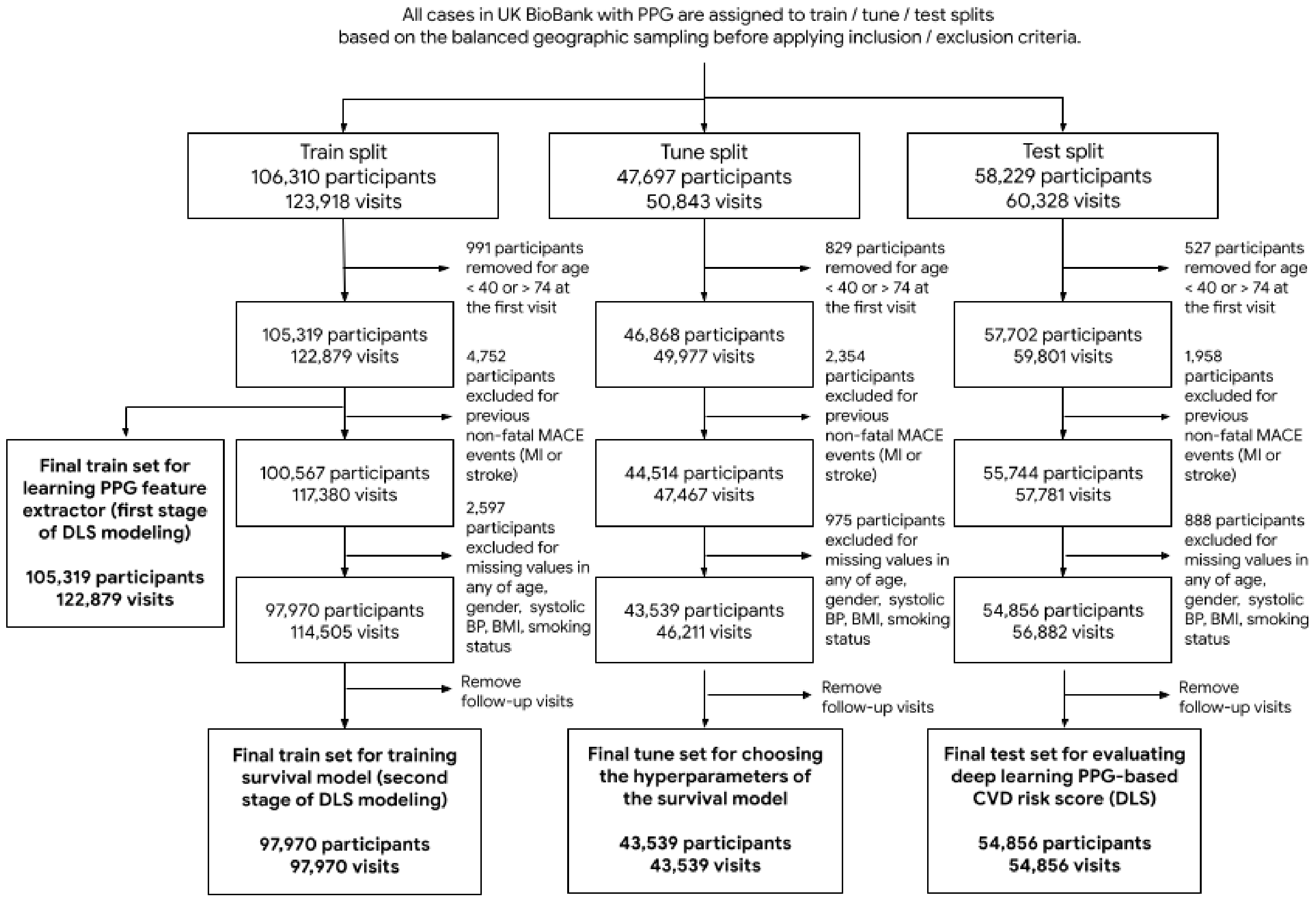

# Supplementary Online Content

## Supplementary Text

### Supplementary Methods

- Details of UK Biobank photoplethysmography data
- Details of model training
- Photoplethysmography morphology-based features

### Supplementary Results

- Analysis for the fixed model operating point at 10% risk
- Models without smoking status
- Applying additional features helps improve the cardiovascular disease risk prediction

## Supplementary Tables

**Supplementary Table 1.** Model performance comparison of 10-year major adverse cardiovascular event (MACE) risk prediction between DLS versus other methods at the 10% risk threshold.
**Supplementary Table 2.** Model performance comparison of 10-year major adverse cardiovascular events (MACE) risk prediction between DLS versus DLS+ (adding BMI) and DLS++ (adding BMI and SBP).
**Supplementary Table 3.** Coefficients and hazard ratios from the Cox's models for 10-year major adverse cardiovascular events (MACE) risk prediction on the UK Biobank (UKB) cohort using DLS, DLS+ and DLS++.
**Supplementary Table 4.** Comparison of 10-year major adverse cardiovascular event (MACE) risk prediction performance between different subgroups using DLS versus office-based refit-WHO model.
**Supplementary Table 5.** UK Biobank variables used in the study.
**Supplementary Table 6.** Training setup for the photoplethysmography (PPG) feature extractor.
**Supplementary Table 7**. Features used in different models for comparison.
**Supplementary Table 8.** The list of proxy tasks used for multitask learning.
**Supplementary Table 9.** Model performance comparison of 10-year major adverse cardiovascular event (MACE) risk prediction between the office-based reference and DLS, and models without smoking status.

## Supplementary Figures

**Supplementary Figure 1.** Overview of our deep learning-based risk prediction model, DLS.
**Supplementary Figure 2.** Kaplan-Meier estimation of DLS with different operating points.
**Supplementary Figure 3.** Calibration plots for all subgroups.
**Supplementary Figure 4.** Prevalence of major adverse cardiovascular event (MACE) in individuals according to model-predicted risk percentiles.

This supplementary material has been provided by the authors to give readers additional information about their work.

# Supplementary Text

## Supplementary Methods

### Details of UK Biobank photoplethysmography data

The photoplethysmography (PPG) waveforms in the UKB (Data field 4205) were acquired using the PulseTrace PCA2 device (CareFusion, USA). The device collected and averaged a minimum of six heart beats per user with a pulse interval close to the average pulse interval. It has been shown that morphological properties of a single representative PPG waveform are related to CV aging and CVD risk, such as augmentation index [1–3]. We preprocessed PPG waveforms by re-scaling to [0, 1].

### Details of model training

During model training, we adopted a multitask learning framework with multiple proxy prediction tasks, such as age, sex, BMI, blood pressure, laboratory data predictions (Supplementary Table 8).

We also used a custom data augmentation inspired by Brownian motion which we termed Brownian tape speed augmentation. We developed the Brownian tape speed augmentation technique, inspired by Brownian motion, to improve the generalizability of the model. The method simulates playing back the signal on a tape while the tape's playback speed is varying according to Brownian motion. Specifically, the playback speed at each time step is drawn from a normal distribution. The method has a single hyperparameter, which we call the magnitude, that is used to define the standard deviation of this normal distribution. For each sequence (i.e., PPG signal), the magnitude is divided by the sequence length to set the standard deviation of the normal distribution for that sequence. This division ensures that regardless of the length of the sequence, the overall amount of transformation is similar.

We then calculate a running sum of this array of normal distribution samples, and add 1 everywhere in order to simulate a random walk of tape speed starting at 1. The array now represents the tape speed. We then calculate another running sum, and now the array represents displacement. We use this displacement as a flow field which is then applied to transform the input using the `tfa_image.dense_image_warp` function in tensorflow.

Training setup for the PPG feature extractor is listed in Supplementary Table 6.

### Photoplethysmography morphology-based features

In the metadata + PPG morphology model, we used the engineered features available in the UK Biobank for PPG-based arterial stiffness evaluation (Supplementary Table 5). The features are pulse wave reflection index (RI), peak to peak time, pulse wave peak position, pulse wave notch position, pulse wave shoulder position, the presence/absence of dicrotic notch, and arterial stiffness index derived from the peak to peak time and the height of the participant.

### Polygenic risk model creation

Individuals of European genetic ancestry who did not have PPG data were split into genome-wide association study (GWAS) (N=208k), train (N=40k), and tune (N=40k) sets. We performed GWAS on the GWAS dataset using BOLT-LMM v2.3.6 [4] and adjusting for age, sex, genotyping array, smoking status, and the top 15 genetic principal components for the following 24 cardiovascular disease-related phenotypes: angina, myocardial infarction, coronary artery disease, heart failure, stroke, cardiovascular death, hypertension, atrial fibrillation, rheumatic heart disease, rheumatoid arthritis, chronic renal failure, diabetes, systolic blood pressure, diastolic blood pressure, low-density lipoprotein (LDL) cholesterol, high-density lipoprotein (HDL) cholesterol, total cholesterol, triglycerides, hemoglobin A1C, glucose, boday mass index (BMI), and three definitions of major adverse cardiovascular event (MACE) as the logical OR of myocardial infarction, cardiovascular death, and stroke; the logical OR of myocardial infarction, cardiovascular death, and heart failure; and the logical OR of myocardial infarction, cardiovascular death, stroke, heart failure, angina, and coronary artery disease (MACE-lenient). For each phenotype, a polygenic risk score (PRS) was generated by BOLT-LMM using the `--predBetasFile` option. Additionally, we ran PolyFun [5] to create functionally-informed fine-mapping PRS. We trained a multilayer perceptron to predict the MACE-lenient phenotype from the 48 PRSs in the train set and selected hyperparameters based on performance in the tune set. We then applied the model to all individuals with PPG data and used the resulting model prediction as the MACE PRS.

## Supplementary Results

### Analysis for the fixed model operating point at 10% risk

At the threshold of 10% risk suggested by the Globorisk study, the sensitivity, specificity, and NRI of DLS was 4.0% (3.0, 5.1), 98.9% (98.8, 99.0), 0.6% (-0.6, 1.7), respectively. Meanwhile, the sensitivity and specificity of the office-based refit-WHO was 3.0% (2.2, 4.0), 99.1% (99.0, 99.2), respectively. We found that without any medical device-dependent measurement, DLS is non-inferior to the office-based refit-WHO score given the risk threshold of 10% suggested by Globorisk study. The full evaluation of all models is listed in the Supplementary Table 1.

### Models without smoking status

We further examined Cox's models without using smoking status as a feature (the office-based refit-WHO score without smoking status, DLS without smoking status). We found that removing smoking status from the predictor set did not reduce the DLS performance and non-inferiority relative to the office-based refit-WHO was maintained (Supplementary Table 9). However, the calibration was worse (slope of 0.968).

### Applying additional features helps improve the cardiovascular disease risk prediction

In Supplementary Table 2, we demonstrated that adding BMI and SBP on top of the DLS model helps improve the model performance on ten-year MACE risk prediction. We also showed that the lab-based refit-WHO model outperformed the office-based model on C-statistic and specificity matching the sensitivity. Meanwhile, we developed a Full model that included most risk factors used in QRISK and/or ASCVD [6,7]—age, ethnicity, deprivation (IMD score), sex, smoking status, BMI, SBP, glucose level, total cholesterol, HDL, medication for hypertension, past medical history of angina, chronic kidney disease, diabetes, erectile dysfunction, mental illness, rheumatoid arthritis, and systemic lupus erythematosus. The Full model was compared with the office-based reference on a smaller cohort subset due to missing data. The Full model yielded the C-statistic of 73.5% (72.3, 74.8), while the DLS got 71.3% (70.0, 72.6), and the office-based model got 71.2% (70.0, 72.5) (superiority test p<0.01 for both) on the same UKB subset. Regarding the specificity matching the sensitivity of 55.2%, the Full model was 76.5% (76.1, 76.9), while the DLS was 74.3% (72.2, 76.6), and the office-based model was 73.7% (73.3, 74.1) (both p<0.01). While matching the specificity of 63.7%, the sensitivity was 71.1% (68.5, 73.6) versus 68.2% (65.5, 70.8) for DLS and 68.8% (66.2, 71.4) for the office-based model (both p<0.01). We concluded that the Full model showed a better MACE prediction on the UKB subset with the variables available for analysis.

In Supplementary Figure 4, we investigated the extent to which individuals predicted to be at high risk by a model are enriched for MACE prevalence. As expected given the observed improvement in C-statistic, the DLS+ model shows superior performance to the Metadata+ model that contains only age, sex, smoking status, and BMI when examining MACE prevalence in the top 5% and 10% of predicted risk. We observed a similar, and slightly more pronounced, improvement from a model that includes a PRS component in addition to the Metadata+ features at the same 5% and 10% most extreme risk percentiles (2.39- and 2.67-fold enrichment over total sample prevalence, respectively, compared to 2.14- and 2.26-fold enrichment for Metadata+). Interestingly, the contributions of PPG and genetic risk appear complementary, as a model that includes Metadata+, PPG, and PRS was most enriched for MACE prevalence (2.53- and 2.87-fold enrichment, respectively).

While both the Full model and the model including polygenic risk show improved MACE prediction performance, we note that each requires more variables that may not be available in low-resource settings, which may limit the use of such lab-based approaches like QRISK, ASCVD, and PRS.

## Supplementary Tables

**Supplementary Table 1: Model performance comparison of 10-year major adverse cardiovascular event (MACE) risk prediction between DLS versus other methods at the 10% risk threshold.** The sensitivity, specificity, and net reclassification improvement (NRI) were calculated at the 10% risk threshold suggested by the Globorisk study for the British population [9]. CIs of sensitivity and specificity were obtained from the Clopper-Pearson exact method, and the p-values were calculated by the permutation test with a prespecified margin of 2.5% and alpha of 0.05. The 95% CIs of NRI were computed by bootstrapping.

| | Sensitivity@risk threshold=0.1 | | | | Specificity@risk threshold=0.1 | | | | | | |
|---|---|---|---|---|---|---|---|---|---|---|---|
| **Model** | **Mean (%)** | **Delta (%)** | **Non-inferiority p-value** | **Superiority p-value** | **Mean (%)** | **Delta (%)** | **Non-inferiority p-value** | **Superiority p-value** | **NRI (%)** | **NRI (event) (%)** | **NRI (non-event) (%)** |
| Office-based refit-WHO | 3.0 (2.2, 4.0) | reference | | | 99.1 (99.0, 99.2) | reference | | | | | |
| DLS | 4.0 (3.0, 5.1) | 1.0 (-0.3, 2.1) | <0.01 | 0.082 | 98.9 (98.8, 99.0) | -0.2 (-0.3, -0.1) | <0.01 | 0.999 | 0.6 (-0.6, 1.7) | 0.8 (-0.3, 1.9) | 0.2 (0.1, 0.3) |
| DLS+ | 7.4 (6.1, 8.9) | 4.4 (3.0, 5.8) | <0.01 | <0.01 | 98.1 (98.0, 98.2) | -1.0 (-1.1, -0.9) | <0.01 | 1 | 3.0 (1.7, 4.3) | 4.0 (2.8, 5.3) | 1.1 (0.9, 1.2) |
| DLS++ | 4.3 (3.3, 5.5) | 1.3 (0.3, 2.3) | <0.01 | <0.01 | 99.0 (98.9, 99.1) | -0.1 (-0.2, -0.0) | <0.01 | 0.998 | 1.2 (0.3, 2.1) | 1.3 (0.4, 2.2) | 0.1 (0.0, 0.2) |
| Metadata | 0.1 (0.0, 0.4) | -3.0 (-3.9, -2.1) | 0.904 | 1 | 100.0 (100.0, 100.0) | 0.9 (0.8, 1.0) | <0.01 | <0.01 | -2.2 (-3.1, -1.4) | -3.1 (-3.9, -2.2) | -0.9 (-0.9, -0.8) |
| Evaluated on the subset with all PPG morphology data available | | | | | | | | | | | |
| Metadata + PPG morphology | 2.2 (1.5, 3.1) | -0.8 (-1.9, 0.2) | <0.01 | 0.977 | 99.5 (99.4, 99.6) | 0.4 (0.3, 0.5) | <0.01 | <0.01 | -0.6 (-1.6, 0.4) | -1.0 (-1.9, 0.0) | -0.4 (-0.5, -0.3) |
| Office-based refit-WHO | 3.0 (2.2, 4.0) | n/a (subset reference) | | | 99.1 (99.0, 99.2) | n/a (subset reference) | | | | | |
| Evaluated on the subset with laboratory data available | | | | | | | | | | | |
| Lab-based refit-WHO | 3.6 (2.7, 4.8) | 0.7 (-0.3, 1.8) | <0.01 | 0.076 | 99.0 (98.9, 99.1) | -0.1 (-0.2, -0.0) | <0.01 | 0.989 | 0.6 (-0.4, 1.6) | 0.7 (-0.3, 1.8) | 0.1 (0.0, 0.2) |
| Office-based refit-WHO | 2.9 (2.1, 3.9) | n/a (subset reference) | | | 99.1 (99.0, 99.2) | n/a (subset reference) | | | | | |

**Supplementary Table 2: Model performance comparison of 10-year major adverse cardiovascular events (MACE) risk prediction between DLS versus DLS+ (adding BMI) and DLS++ (adding BMI and SBP).** (a) We examined the discrimination performance using C-statistic, reclassification improvement using category-free net reclassification improvement (cfNRI), and model calibration using the slope value from the reliability diagram. *In "Feature used" column, "Metadata" includes age, sex, and smoking status. (b) The sensitivity was calculated at the risk threshold matching the specificity of SBP-140, and the specificity was calculated at the risk threshold matching the sensitivity of SBP-140. 95% confidence intervals (CIs) of C-statistic, cfNRI, and slope were obtained from the bootstrapping, and p-values were computed by the permutation test. CIs of sensitivity and specificity were obtained from the Clopper-Pearson exact method, and the p-values were calculated by a permutation test with the prespecified margin of 2.5% and alpha of 0.05. The 95% CIs of NRI were computed by bootstrapping.

(a)

| Model | C-statistic (%) | Delta in C-statistic (%) | P-value for non-inferiority of C-statistic | P-value for superiority of C-statistic | cfNRI (%) | cfNRI (event) (%) | cfNRI (non-event) (%) | Calibration slope | Features used* |
|---|---|---|---|---|---|---|---|---|---|
| Office-based refit-WHO | 70.9 (69.7, 72.2) | n/a (reference) | | | | | | 0.979 (0.915, 1.038) | Metadata + BMI + SBP |
| DLS | 71.1 (69.9, 72.4) | 0.2 (-0.4, 0.8) | <0.01 | 0.292 | 0.1 (-0.0, 0.1) | 0.1 (-0.0, 0.2) | 0.0 (0.0, 0.0) | 0.981 (0.919, 1.045) | Metadata + PPG |
| DLS+ | 71.3 (70.2, 72.7) | 0.5 (-0.1, 1.0) | <0.01 | 0.073 | 0.3 (0.2, 0.4) | 0.4 (0.3, 0.5) | 0.1 (0.1, 0.1) | 1.079 (1.001, 1.148) | Metadata + BMI + PPG |
| DLS++ | 71.9 (70.8, 73.2) | 1.0 (0.6, 1.4) | <0.01 | <0.01 | 0.2 (0.1, 0.2) | 0.2 (0.1, 0.2) | -0.0 (-0.0, -0.0) | 0.952 (0.89, 1.01) | Metadata + BMI + SBP + PPG |

(b)

| Model | Sensitivity@specificity of 63.7% | | | | | | | Specificity@sensitivity of 55.2% | | | | | | |
|---|---|---|---|---|---|---|---|---|---|---|---|---|---|---|
| | Mean (%) | Delta (%) | Non-inferiority p-value | Superiority p-value | NRI (%) | NRI (event) (%) | NRI (non-event) (%) | Mean (%) | Delta (%) | Non-inferiority p-value | Superiority p-value | NRI (%) | NRI (event) (%) | NRI (non-event) (%) |
| Office-based refit-WHO | 67.7 (65.2, 70.1) | reference | | | | | | 73.1 (72.7, 73.5) | reference | | | | | |

| Model | Sensitivity@specificity of 63.7% | | | | | | | Specificity@sensitivity of 55.2% | | | | | | |
|---|---|---|---|---|---|---|---|---|---|---|---|---|---|---|
| | Mean (%) | Delta (%) | Non-inferiority p-value | Superiority p-value | NRI (%) | NRI (event) (%) | NRI (non-event) (%) | Mean (%) | Delta (%) | Non-inferiority p-value | Superiority p-value | NRI (%) | NRI (event) (%) | NRI (non-event) (%) |
| DLS | 67.9 (65.4, 70.3) | 0.1 (-1.9, 2.0) | 0.012 | 0.654 | -0.3 (-2.0, 1.6) | 1.0 (-0.9, 2.9) | 1.2 (0.9, 1.5) | 74.0 (73.6, 74.4) | 0.9 (-0.7, 2.5) | <0.01 | <0.01 | 1.1 (-0.9, 3.1) | 1.6 (-0.5, 3.4) | 0.4 (0.1, 0.8) |
| DLS+ | 67.9 (65.4, 70.3) | 0.1 (-1.9, 2.2) | <0.01 | 0.5 | 0.0 (-1.9, 2.0) | 0.6 (-1.2, 2.6) | 0.5 (0.2, 0.8) | 74.7 (74.3, 75.0) | 1.4 (-0.2, 2.8) | <0.01 | <0.01 | 2.4 (0.3, 4.3) | 3.5 (1.4, 5.4) | 1.1 (0.8, 1.4) |
| DLS++ | 68.8 (66.3, 71.2) | 1.1 (-0.4, 2.6) | <0.01 | 0.086 | 1.1 (-0.4, 2.5) | 1.3 (-0.1, 2.7) | 0.3 (-0.0, 0.5) | 75.2 (74.8, 75.5) | 2.0 (0.9, 3.1) | <0.01 | <0.01 | 2.6 (1.1, 4.2) | 2.6 (1.1, 4.2) | -0.0 (-0.2, 0.2) |

**Supplementary Table 3: Coefficients and hazard ratios from the Cox's models for 10-year major adverse cardiovascular events (MACE) risk prediction on the UK Biobank (UKB) cohort using DLS, DLS+ and DLS++.** Hazard ratios are shown at the median age of the MACE event, which is 63 years in the train split of UKB cohort. Hazard ratios for smokers are for men, and their interaction with sex shows the adjusted risk for women. We included interaction terms between age and other predictors because the HRs for proportional effects on CVD declined with age [10,11].

| | **DLS** | | | | **DLS+** | | | | **DLS++** | | | | **Office-based refit-WHO** | | |
|---|---|---|---|---|---|---|---|---|---|---|---|---|---|---|---|
| **Predictor** | **Main effect** | **Age interaction term** | **Hazard Ratio** | **PPG feature ranges (mean, SD)** | **Main effect** | **Age interaction term** | **Hazard Ratio** | **PPG feature ranges (mean, SD)** | **Main effect** | **Age interaction term** | **Hazard Ratio** | **PPG feature ranges (mean, SD)** | **Main effect** | **Age interaction term** | **Hazard Ratio** |
| Male Smoker | 0.581 (p=0.094) | -0.0462 (p=0.407) | 1.337 (1.257, 1.422) | - | 0.652 (p=0.061) | -0.0608 (p=0.276) | 1.309 (1.239, 1.382) | - | 0.624 (p=0.075) | -0.0548 (p=0.329) | 1.322 (1.254, 1.395) | - | 0.334 (p=0.061) | -0.0069 (p=0.809) | 1.337 (1.306, 1.37) |
| Female smoker | -0.02309 (p=0.788) | - | 0.977 (0.919, 1.039) | - | -0.02905 (p=0.736) | | 0.971 (0.92, 1.026) | - | -0.0387 (p=0.653) | - | 0.962 (0.912, 1.015) | - | 0.02069 (p=0.78) | - | 1.021 (0.997, 1.046) |
| Body mass index | - | - | - | - | 0.029 (p=0.38) | 0.0017 (p=0.749) | 1.041 (0.996, 1.087) | - | 0.004 (p=0.896) | 0.0047 (p=0.402) | 1.034 (0.994, 1.076) | - | 0.007 (p=0.649) | 0.0048 (p=0.059) | 1.038 (1.02, 1.056) |
| Systolic blood pressure | - | - | - | - | - | - | - | - | 0.022 (p=0.014) | -0.0025 (p=0.079) | 1.006 (0.982, 1.031) | - | 0.007 (p=0.037) | 0.0002 (p=0.682) | 1.009 (1.0, 1.018) |
| PPG features | | | | | | | | | | | | | | | |
| PPG-1 | -0.111 (p=0.0) | - | 0.895 (0.879, 0.912) | 0.245 (3.17) | -0.113 (p=0.0) | - | 0.893 (0.874, 0.913) | 0.184 (2.725) | -0.094 (p=0.0) | - | 0.91 (0.893, 0.928) | 0.253 (3.125) | - | | |

| Predictor | DLS | | | | DLS+ | | | | DLS++ | | | |
|---|---|---|---|---|---|---|---|---|---|---|---|---|
| | **Main effect** | **Age interaction term** | **Hazard Ratio** | **PPG feature ranges (mean, SD)** | **Main effect** | **Age interaction term** | **Hazard Ratio** | **PPG feature ranges (mean, SD)** | **Main effect** | **Age interaction term** | **Hazard Ratio** | **PPG feature ranges (mean, SD)** |
| PPG-2 | 0.059 (p=0.002) | - | 1.061 (1.021, 1.101) | -0.047 (1.195) | 0.073 (p=0.001) | - | 1.076 (1.031, 1.123) | -0.014 (0.978) | 0.011 (p=0.638) | - | 1.011 (0.966, 1.058) | -0.004 (1.092) |
| PPG-3 | 0.002 (p=0.915) | - | 1.002 (0.965, 1.041) | -0.041 (1.071) | 0.014 (p=0.574) | - | 1.014 (0.966, 1.064) | -0.01 (0.881) | 0.052 (p=0.035) | - | 1.053 (1.004, 1.105) | -0.004 (0.947) |
| PPG-4 | 0.032 (p=0.128) | - | 1.033 (0.991, 1.077) | 0.022 (1.016) | 0.133 (p=0.0) | - | 1.142 (1.082, 1.205) | -0.005 (0.791) | -0.001 (p=0.956) | - | 0.999 (0.953, 1.047) | -0.034 (0.915) |
| PPG-5 | -0.055 (p=0.025) | - | 0.947 (0.902, 0.993) | -0.06 (0.894) | 0.019 (p=0.516) | - | 1.019 (0.962, 1.079) | -0.026 (0.697) | -0.014 (p=0.577) | - | 0.986 (0.937, 1.037) | -0.06 (0.875) |
| PPG-Heart Rate | 0.006 (p=0.001) | - | 1.006 (1.002, 1.009) | - | 0.003 (p=0.124) | - | 1.003 (0.999, 1.006) | - | 0.002 (p=0.245) | - | 1.002 (0.998, 1.006) | - |

**Supplementary Table 4: Comparison of 10-year major adverse cardiovascular event (MACE) risk prediction performance between different subgroups using DLS versus office-based refit-WHO model.** The sensitivity and specificity were calculated at the risk threshold matching SBP-140's specificity (see Statistical Analysis). 95% confidence intervals (CIs) were obtained from the Clopper-Pearson exact method.

| Subgroup | Model | C-statistic | Non-inferiority p-value | Superiority p-value | Sensitivity | Specificity | Average predicted risk score | Slope |
|---|---|---|---|---|---|---|---|---|
| Never smoked | Office-based refit-WHO | 72.1 (70.6, 73.9) | - | - | 0.598 (0.562, 0.632) | 0.722 (0.716, 0.727) | 0.023 | 0.73 |
| | DLS | 71.7 (70.1, 73.3) | 0.11 | 0.815 | 0.558 (0.522, 0.593) | 0.731 (0.726, 0.736) | 0.023 | 0.76 |
| Smoked | Office-based refit-WHO | 68.9 (67.5, 70.4) | - | - | 0.756 (0.725, 0.785) | 0.51 (0.504, 0.517) | 0.035 | 1.01 |
| | DLS | 69.9 (68.2, 71.3) | <0.01 | 0.029 | 0.786 (0.757, 0.814) | 0.498 (0.491, 0.504) | 0.036 | 0.96 |
| Age<55 | Office-based refit-WHO | 68.7 (66.0, 71.5) | - | - | 0.157 (0.117, 0.205) | 0.936 (0.933, 0.939) | 0.013 | 0.77 |
| | DLS | 69.1 (66.6, 72.4) | 0.088 | 0.335 | 0.22 (0.174, 0.273) | 0.93 (0.926, 0.933) | 0.014 | 0.86 |
| Age>=55 | Office-based refit-WHO | 65.4 (64.1, 66.8) | - | - | 0.793 (0.77, 0.815) | 0.426 (0.42, 0.431) | 0.038 | 0.95 |
| | DLS | 65.6 (64.2, 66.8) | <0.01 | 0.335 | 0.775 (0.751, 0.797) | 0.429 (0.424, 0.435) | 0.038 | 0.95 |
| Female | Office-based refit-WHO | 70.8 (68.8, 72.4) | - | - | 0.422 (0.382, 0.463) | 0.82 (0.815, 0.824) | 0.018 | 0.70 |
| | DLS | 69.5 (68.0, 71.2) | 0.578 | 0.958 | 0.406 (0.366, 0.446) | 0.804 (0.8, 0.809) | 0.019 | 0.78 |
| Male | Office-based refit-WHO | 66.0 (64.5, 67.6) | - | - | 0.836 (0.811, 0.858) | 0.377 (0.37, 0.383) | 0.041 | 0.99 |
| | DLS | 67.4 (66.0, 68.9) | <0.01 | <0.01 | 0.84 (0.815, | 0.396 (0.39, 0.403) | 0.041 | 0.95 |
| | | | | | 0.862) | | | |
| **Subgroup** | **Model** | **C-statistic** | **Non-inferiority p-value** | **Superiority p-value** | **Sensitivity** | **Specificity** | **Average predicted risk score** | **Slope** |
| HbA1c <=48 | Office-based refit-WHO | 71.0 (69.8, 72.2) | - | - | 0.677 (0.652, 0.701) | 0.636 (0.632, 0.64) | 0.028 | 0.96 |
| | DLS | 71.2 (69.9, 72.4) | <0.01 | 0.37 | 0.671 (0.646, 0.696) | 0.635 (0.631, 0.639) | 0.028 | 0.95 |
| HbA1c >48 | Office-based refit-WHO | 59.2 (55.7, 62.8) | - | - | 0.701 (0.629, 0.766) | 0.43 (0.407, 0.453) | 0.039 | 0.69 |
| | DLS | 60.6 (56.6, 64.0) | 0.03 | 0.148 | 0.712 (0.641, 0.776) | 0.447 (0.424, 0.47) | 0.04 | 0.66 |
| No hypertension | Office-based refit-WHO | 70.8 (69.3, 72.4) | - | - | 0.589 (0.556, 0.623) | 0.703 (0.698, 0.707) | 0.024 | 1.03 |
| | DLS | 71.2 (69.6, 72.9) | <0.01 | 0.134 | 0.613 (0.579, 0.646) | 0.686 (0.681, 0.69) | 0.025 | 1.08 |
| Hypertension | Office-based refit-WHO | 64.4 (62.5, 66.2) | - | - | 0.783 (0.751, 0.812) | 0.42 (0.411, 0.428) | 0.039 | 0.9 |
| | DLS | 65.3 (63.6, 67.1) | <0.01 | 0.046 | 0.748 (0.715, 0.779) | 0.466 (0.458, 0.475) | 0.037 | 0.84 |

**Supplementary Table 5: UK Biobank variables used in the study.**

| Variable | UK Biobank data field | Notes |
|---|---|---|
| UKB site | 54 | For data split |
| Visit date | 53 | To identify the age at visit |
| Age | 21003 | |
| Sex | 31 | |
| Smoking status | 20116 | |
| Ethnicity | 21000 | For Full model |
| Deprivation | 26410, 26426, 26427 | For Full model |
| BMI | 21001 | |
| SBP | 4080 | |
| Glucose | 30740 | For lab-based score |
| Total cholesterol | 30690 | For lab-based score |
| HbA1c | 30750 | For subgroup analysis |
| HDL | 30760 | For Full model |
| LDL | 30780 | For Full model |
| PPG waveform | 4205 | For all DLS scores |
| PPG pulse rate | 4194 | For all DLS scores and the score using engineered PPG morphology |
| PPG reflection index | 4195 | For the score using engineered PPG morphology |
| PPG peak-to-peak time | 4196 | For the score using engineered PPG morphology |
| PPG peak position | 4198 | For the score using engineered PPG morphology |
| PPG notch position | 4199 | For the score using engineered PPG morphology |
| PPG shoulder position | 4200 | For the score using engineered PPG morphology |
| PPG notch absent | 4204 | For the score using engineered PPG morphology |
| PPG ASI | 21021 | For the score using engineered PPG morphology |
| Hypertension | 6150 (self-reported), 131286 (ICD), 131288 (ICD), 131290 (ICD), 131292 (ICD) | For subgroup analysis |
| Myocardial infarction | 3894 (age at diagnosis), 6150 (self-reported), 42000 (self-reported), 131298 (ICD), 131300 (ICD), 131302 (ICD) | MACE outcome |
| Stroke | 4056 (age at diagnosis), 6150 (self-reported), 131368 (ICD), 131366 (ICD), 42006 (self-reported/EHR), 42008 (self-reported/EHR) | MACE outcome |
| Cardiovascular-related death | 40000 (ICD), 40001 (ICD), 40010 (text: cause of death) | MACE outcome, we considered all cause of death related to coronary artery diseases, myocardial infarction, stroke, hypertensive problems (including ICH, SAH, AAA, etc.), and other heart-related problems |
| Medication | 20003 (self-reported for all medications), 6177 (for statin, hypertension medication, insulin) | For Full model |
| Medical conditions | 20002 (self-reported) | For the Full model, including all self-reported medical history information, e.g. angina, heart failure, hyperlipidemia, erectile dysfunction, mental illness, migraine, chronic kidney disease, rheumatoid arthritis, SLE, etc. |

**Supplementary Table 6: Training setup for the photoplethysmography (PPG) feature extractor.**

| **Hyperparameter** | **DLS** | **DLS+** | **DLS++** |
|---|---|---|---|
| Neural network architecture | ResNet-18 | | |
| Dropout rate | 0.0 | | |
| Epochs | 80 | | |
| Optimizer | AdamW [8] | | |
| Learning rate | 0.0001 with cosine 1-epoch warmup | 0.0003 with cosine 1-epoch warmup | 0.0001 with cosine 1-epoch warmup |
| Weight decay | 0.000003 | 0.0001 | 0.000003 |
| Augmentation | Brownian tape speed with magnitude of 2 and randomly applied 50% (1/2) of the time | Brownian tape speed with magnitude of 0.1 and randomly applied 50% (1/2) of the time | Brownian tape speed with magnitude of 2 and randomly applied 50% (1/2) of the time |
| Ridge penalization parameter for the Cox model | 0.00003 | 0.00003 | 0.00003 |

**Supplementary Table 7: Features used in different models for comparison.** We compared all methods with the office-based refit-WHO model. The evaluations of DLS models and additional reference methods are in the main content Supplementary Tables 1, 2, 9. For the supplementary reference methods, the results are listed in the Supplementary Tables. *Lab-based refit-WHO and metadata + PPG morphology models are compared with a subset of the whole cohort. **The full model used most QRISK features. The detail of the feature set is described in the Supplementary Methods.

| Method | Age | Sex | Smoking status | BMI | SBP | Labs | Medical history / family history / medications | PPG (engineered) | PPG (deep learning-based) |
|---|---|---|---|---|---|---|---|---|---|
| DLS | x | x | x | | | | | | x |
| DLS+ | x | x | x | x | | | | | x |
| DLS++ | x | x | x | x | x | | | | x |
| Main reference method | | | | | | | | | |
| Office-based refit-WHO | x | x | x | x | x | | | | |
| Additional reference methods | | | | | | | | | |
| Metadata | x | x | x | | | | | | |
| Lab-based refit-WHO* | x | x | x | | x | x | | | |
| Metadata + PPG morphology* | x | x | x | | | | | x | |
| Supplementary reference methods | | | | | | | | | |
| Smoking status-only | | | x | | | | | | |
| Office without smoking status | x | x | | x | x | | | | |
| DLS without smoking status | x | x | | | | | | | x |
| Full** | x | x | x | x | x | x | x | | |

**Supplementary Table 8: The list of proxy tasks used for multitask learning.**

| Variable | Type of task |
|---|---|
| Sex | Classification |
| Chronological age | Regression |
| Body mass index (thresholded at 33 $kg/m^2$) | Classification |
| Hypertension status | Classification |
| HbA1c (thresholded at 48 mmol/mol / 6.5%) | Classification |
| Total cholesterol (thresholded at 7.16 mmol/L) | Classification |
| Systolic blood pressure (thresholded at 160 mmHg) | Classification |
| Previous MACE event | Classification |
| PPG dicrotic notch | Classification |

**Supplementary Table 9: Model performance comparison of 10-year major adverse cardiovascular event (MACE) risk prediction between the office-based reference and DLS, and models without smoking status.** (a) We examined the ability of discrimination using C-statistic, reclassification improvement using category-free net reclassification improvement (cfNRI), and model calibration using the slope value from the reliability diagram. *In “Feature used” column, “Metadata” includes age, sex, and smoking status. (b) The sensitivity was calculated at the risk threshold matching specificity of the SBP-140 baseline at 63.7%, and the specificity was calculated based on the risk threshold matching sensitivity of the SBP-140 baseline at 55.2%. 95% confidence intervals (CIs) of C-statistic, cfNRI, and slope were obtained from the bootstrapping, and the p-values were computed by the permutation test. CIs of sensitivity and specificity were obtained from the Clopper-Pearson exact method, and the p-values were calculated by the permutation test with the prespecified margin of 2.5% and alpha of 0.05. The 95% CIs of NRI were computed by bootstrapping.

(a)

| **Model** | **C-statistic (%)** | **Delta (%)** | **P-value for non-inferiority of C-statistic** | **P-value for superiority of C-statistic** | **cfNRI (%)** | **cfNRI (event) (%)** | **cfNRI (non-event) (%)** | **Calibration slope** | **Features used**** |
|---|---|---|---|---|---|---|---|---|---|
| Office-based refit-WHO | 70.9 (69.7, 72.2) | n/a (reference) | | | | | | 0.979 (0.915, 1.038) | Metadata + BMI + SBP |
| DLS | 71.1 (69.9, 72.4) | 0.2 (-0.4, 0.8) | 0.001 | 0.292 | 0.1 (-0.0, 0.1) | 0.1 (-0.0, 0.2) | 0.0 (0.0, 0.0) | 0.981 (0.919, 1.045) | Metadata + PPG |
| Smoking status-only | 53.9 (52.6, 55.1) | -17.1 (-18.4, -15.5) | 1 | 1 | -1.5 (-1.6, -1.4) | -1.1 (-1.2, -1.0) | 0.4 (0.4, 0.4) | 1.506 (1.153, 2.204) | Smoking |
| Office without smoking status | 70.8 (69.6, 72.1) | -0.1 (-0.3, 0.2) | 0.001 | 0.667 | -0.0 (-0.1, 0.0) | -0.0 (-0.1, 0.0) | -0.0 (-0.0, -0.0) | 0.982 (0.916, 1.051) | Age, sex, BMI, SBP |
| DLS without smoking status | 71.1 (69.9, 72.4) | 0.2 (-0.4, 0.9) | 0.001 | 0.261 | 0.1 (-0.0, 0.2) | 0.1 (-0.0, 0.2) | 0.0 (0.0, 0.0) | 0.968 (0.901, 1.032) | DLS, age, sex |

(b)

| | **Sensitivity@specificity of 63.7%** | | | | | | | **Specificity@sensitivity of 55.2%** | | | | | | |
|---|---|---|---|---|---|---|---|---|---|---|---|---|---|---|
| Model | Mean (%) | Delta (%) | Non-inferiority p-value | Superiority p-value | NRI (%) | NRI (event) (%) | NRI (non-event) (%) | Mean (%) | Delta (%) | Non-inferiority p-value | Superiority p-value | NRI (%) | NRI (event) (%) | NRI (non-event) (%) |
| Office-based refit-WHO | 67.7 (65.2, 70.1) | reference | | | | | | 73.1 (72.7, 73.5) | reference | | | | | |
| DLS | 67.9 (65.4, 70.3) | 0.1 (-1.9, 2.0) | 0.012 | 0.654 | -0.3 (-2.0, 1.6) | 1.0 (-0.9, 2.9) | 1.2 (0.9, 1.5) | 74.0 (73.6, 74.4) | 0.9 (-0.7, 2.5) | <0.01 | <0.01 | 1.1 (-0.9, 3.1) | 1.6 (-0.5, 3.4) | 0.4 (0.1, 0.8) |
| Office without smoking status | 67.6 (65.2, 70.0) | -0.1 (-1.4, 1.1) | <0.01 | 0.5 | 0.3 (-1.1, 1.7) | 1.9 (0.6, 3.3) | 1.7 (1.4, 1.9) | 74.0 (73.7, 74.4) | 0.7 (-0.5, 1.9) | <0.01 | <0.01 | 1.6 (-0.0, 3.1) | 3.1 (1.5, 4.7) | 1.5 (1.3, 1.8) |
| DLS without smoking status | 68.8 (66.3, 71.2) | 1.1 (-0.9, 3.2) | <0.01 | 0.244 | 0.8 (-1.2, 2.8) | 2.0 (-0.1, 3.9) | 1.1 (0.8, 1.4) | 74.2 (73.8, 74.5) | 1.0 (-0.4, 2.5) | <0.01 | <0.01 | 1.2 (-0.7, 3.1) | 1.6 (-0.4, 3.4) | 0.4 (0.0, 0.7) |

## Supplementary Figures

**Supplementary Figure 1: Overview of our deep learning-based risk prediction model, DLS.** Blue: models; yellow: inputs; red: intermediate data representations (embeddings) obtained from the deep learning-based PPG feature extractor.

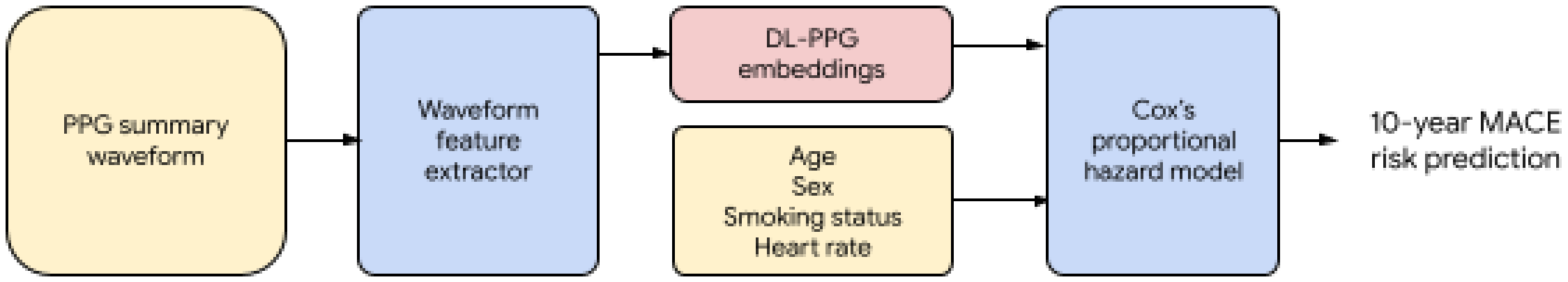

**Supplementary Figure 2: Kaplan-Meier estimation of DLS with different operating points.** We compared the survival estimation between the high and low risk groups, which were defined by the risk threshold at 10% suggested by the Globorisk study [9]. For example, a case with prediction value higher than 0.1 will be high risk, else low risk. The p-values were calculated by the log-rank test.

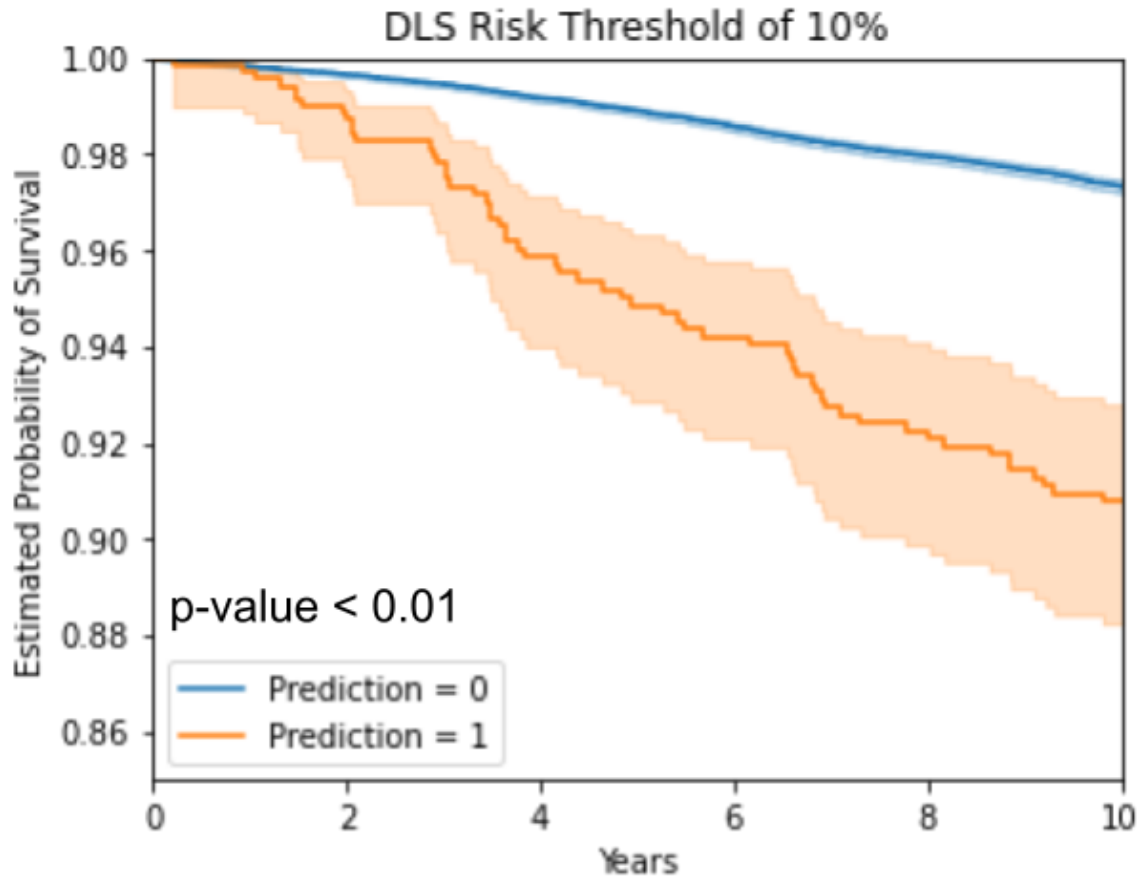

| KM_estimate | 0 | 2 | 4 | 6 | 8 | 10 |
|---|---|---|---|---|---|---|
| At risk | 54152 | 53708 | 51735 | 51425 | 51110 | 50780 |
| Censored | 0 | 263 | 1978 | 1978 | 1978 | 1978 |
| Events | 0 | 181 | 439 | 749 | 1064 | 1394 |

| KM_estimate | 0 | 2 | 4 | 6 | 8 | 10 |
|---|---|---|---|---|---|---|
| At risk | 704 | 685 | 587 | 577 | 565 | 556 |
| Censored | 0 | 10 | 90 | 90 | 90 | 90 |
| Events | 0 | 9 | 27 | 37 | 49 | 58 |

**Supplementary Figure 3: Calibration plots for all subgroups.** The calibration slope values indicate the coefficient of a linear regression where the dependent variable was the fraction of positives (predicted risk) and the independent variable was the mean prediction. We used ten bins to discretize the prediction interval and chose deciles of predicted risk to define the widths of the bins. For the elevated HbA1c subgroup, we used quintiles to ensure sufficient events. All models (office-based refit-WHO, DLS) are calibrated better in smoking, older, male, non A1c elevated, and non-hypertensive subgroups.

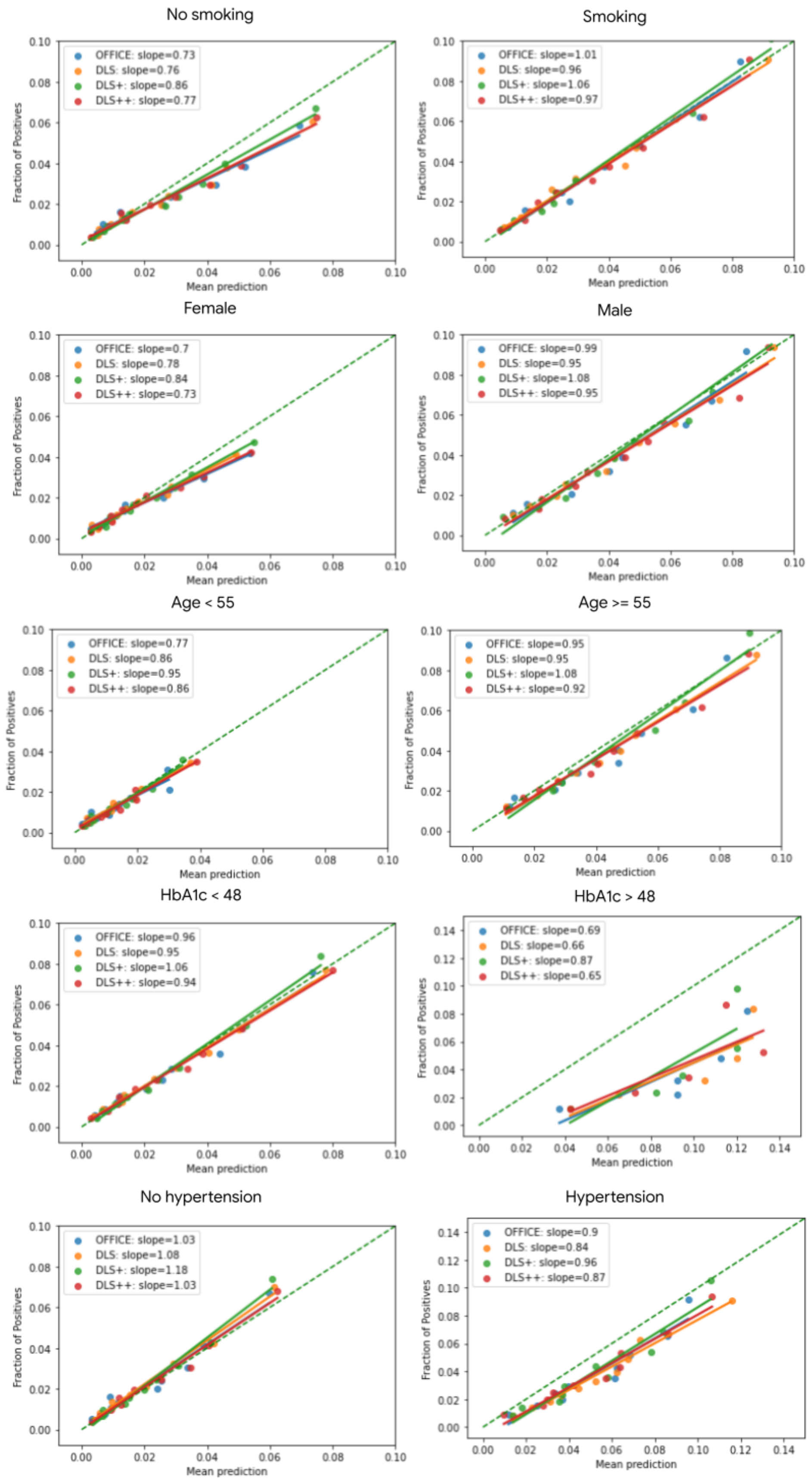

**Supplementary Figure 4. Prevalence of major adverse cardiovascular event (MACE) in individuals according to model-predicted risk percentiles.** For each of four risk models, the prevalence of MACE was computed in the individuals scoring in the highest 20, 10, and 5% risk according to the model. Error bars computed via 100 bootstrap iterations. The dashed gray line shows MACE prevalence in the entire sample. Metadata+, model containing age, sex, smoking status, and BMI. DLS+, model containing age, sex, smoking status, BMI, and PPG. Metadata+ + polygenic risk score (PRS), model containing age, sex, smoking status, BMI, and polygenic risk score. DLS+ + PRS, model containing age, sex, smoking status, BMI, PPG, and PRS.

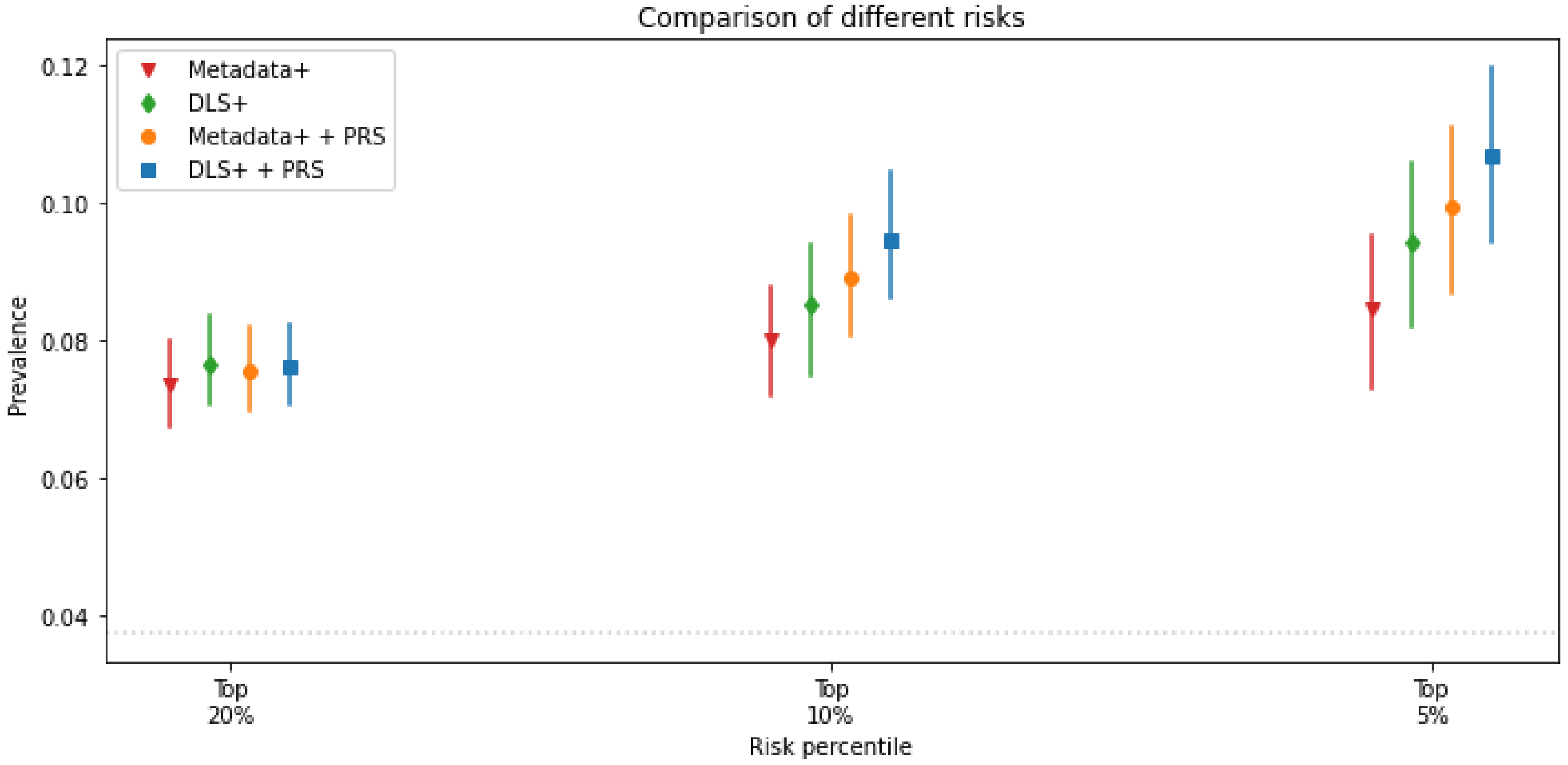

## Supplementary References